\documentclass[letterpaper, 10 pt, conference]{ieeeconf}  

\IEEEoverridecommandlockouts                              

\usepackage{graphicx}
\usepackage{hyperref}
\usepackage{url}
\usepackage{amsmath,amssymb,amsfonts}
\usepackage{algorithmic}
\usepackage{graphicx}
\usepackage{textcomp}
\usepackage{xcolor}
\usepackage{multirow}
\usepackage{booktabs}
\usepackage{subcaption}
\usepackage{array}
\usepackage{tcolorbox}

\title{\LARGE \bf
SkeySpot: Automating Service Key Detection for Digital Electrical Layout Plans in the Construction Industry
}

\author{Dhruv Dosi$^{1}$, Rohit Meena$^{1}$, Param Rajpura$^{1}$, Yogesh Kumar Meena$^{*1}$
\thanks{$^{1}$Human-AI Interaction (HAIx) Lab, IIT Gandhinagar, India
        {\tt\small yk.meena@iitgn.ac.in}}%
}

\begin{document}

\maketitle

\begin{abstract}
Legacy floor plans, often preserved only as scanned documents, remain essential resources for architecture, urban planning, and facility management in the construction industry. However, the lack of machine-readable floor plans render large-scale interpretation both time-consuming and error-prone. Automated symbol spotting offers a scalable solution by enabling the identification of service key symbols directly from floor plans, supporting workflows such as cost estimation, infrastructure maintenance, and regulatory compliance. This work introduces a labelled Digitised Electrical Layout Plans (DELP) dataset comprising 45 scanned electrical layout plans annotated with 2,450 instances across 34 distinct service key classes. A systematic evaluation framework is proposed using pretrained object detection models for DELP dataset. Among the models benchmarked, YOLOv8 achieves the highest performance with a mean Average Precision (mAP) of 82.5\%. Using YOLOv8, we develop SkeySpot, a lightweight, open-source toolkit for real-time detection, classification, and quantification of electrical symbols. SkeySpot produces structured, standardised outputs that can be scaled up for interoperable building information workflows, ultimately enabling compatibility across downstream applications and regulatory platforms. By lowering dependency on proprietary CAD systems and reducing manual annotation effort, this approach makes the digitisation of electrical layouts more accessible to small and medium-sized enterprises (SMEs) in the construction industry, while supporting broader goals of standardisation, interoperability, and sustainability in the built environment.
\end{abstract}

\section{INTRODUCTION}

Digital electrical layout plans are central to construction and facilities management workflows, representing the design and placement of electrical and heating, ventilation, and air conditioning (HVAC) components within construction environments \cite{pizarro2022automatic}. These plans encode spatial and architectural information using standardised graphical symbols, referred as service keys, to represent components like sockets, switches, outlets, fixtures, etc. Generated via Computer-Aided Design (CAD) software, these diagrams are used to estimate costs, plan installations, coordinate maintenance and assess compliance. However, in practice, CAD files are often inaccessible to many stakeholders, particularly small and medium-sized enterprises (SMEs) in the construction industry, who instead rely on scanned floor plans or PDFs. The high cost and technical complexity of tools like AutoCAD \footnote{\href{https://www.autodesk.com/in/solutions/floor-plan} {AutoCAD Floor Plan Software}} further limit their adoption in such contexts, creating a significant gap in accessible, automated annotation solutions.



Manual detection and annotation of symbols in scanned plans remain the prevailing method but are slow, error-prone, and require domain expertise \cite{pizarro2022automatic}. The task is further complicated by densely cluttered diagrams, visual occlusions, and subtle inter-class variations in symbol shapes, sizes, and orientations. These challenges motivate the need for robust, scalable, and easy-to-use automation tools that do not depend on proprietary design software.

\begin{figure*}[htbp!]
  \centering
  \includegraphics[width=.83\textwidth]{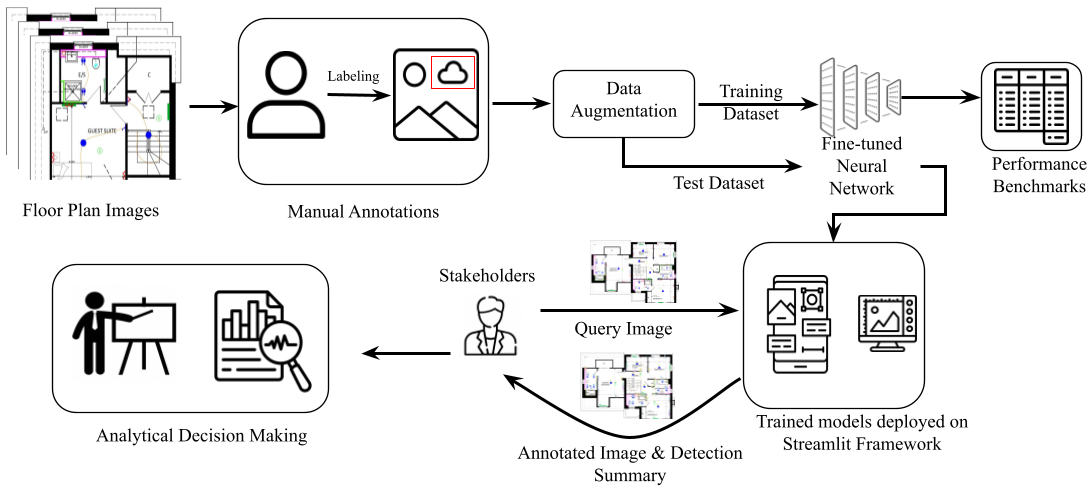}
\caption{Proposed SkeySpot pipeline, outlining the workflow from dataset preparation and annotation to training, benchmarking, and deployment using Streamlit, an open-source Python framework}
\label{fig:Pipeline Diagram}

\end{figure*}

Symbol spotting in architectural diagrams has been an open research topic over the past decade, especially in the context of scanned or rasterised floor plans \cite{pizarro2022automatic}. To the best of our knowledge, there have not been significant contributions toward symbol detection and its real-time deployment for electrical layout plans. To address automatic symbol detection, prior non-deep learning approaches for symbol spotting have primarily relied on two methodological paradigms: pixel-based and vector-based techniques. For instance, \cite{rezvanifar2019symbol} proposes a two-phase approach using low-level descriptors and template matching. Pixel-based methods such as F-signature \cite{tabbone2003matching}, Blurred Shape Model (BSM) \cite{escalera2009blurred}, and Circular BSM (CBSM) \cite{escalera2010circular} operate on raster data, while vector-based approaches like SCIP and ESCIP \cite{nguyen2009symbol} exploit vectorial representations. Structural methods leveraging graphs—such as Attributed Relational Graphs \cite{qureshi2008spotting}, Region Adjacency Graphs \cite{6460131}, graph serialization \cite{dutta2013symbol}, graph embedding \cite{luqman2011subgraph}, and relational indexing \cite{rusinol2010relational}, attempt to capture the topological structure of symbols. Other works have used connected components, constituent primitives, and contour maps \cite{nayef2013use}.


Despite their algorithmic sophistication, these approaches remain sensitive to noise, symbol distortion, and layout heterogeneity. Further, there are multiple symbols that are very similar in color and shape, that are not efficiently classified by rule-based methods \cite{de2014statistical}. In recent years, deep learning-based methods have shown notable progress in digital document analysis. Dey et al. \cite{dey2017shallow} employ a simple LeNet for handwritten symbol recognition, while Riba et al. \cite{riba2017graph} utilize message-passing neural networks to operate on structural graphs. Semantic segmentation models have been applied to architectural diagrams for wall and door detection \cite{yang2018semantic, 8892918}. GAN-based methods have also been explored for symbol spotting \cite{elyan2020deep}. However, many of these approaches are tested on simplified datasets and are not designed for real-time deployment in high-clutter environments. Notably, \cite{8978054, ziran2018object} utilize public datasets of about 1300 images, but these lack the symbol diversity and visual complexity encountered in practical settings. Additionally, previous works \cite{10007281, 10007400} also focus on piping and instrumentation diagrams using Faster R-CNN, which involves slower inference and is not optimised for deployment. Rezvanifar et al. \cite{9151066} address only 12 architectural symbols and do not generalize to more complex electrical layouts.

By combining dataset creation, model evaluation, and deployment in an accessible toolkit, this work fills a significant gap in automated symbol spotting for scanned electrical layouts and offers a robust baseline for further research in the automatic symbol spotting domain.
The main contributions of this work are summarised as follows:

\begin{enumerate}
    \item Introduced a novel dataset of 45 annotated floorplans, covering 34 unique service key classes and a total of 2450 instances across the service keys. 
    \item Benchmarked Faster R-CNN and YOLOv8 models fine-tuned for the service key detection and classification, offering a comprehensive performance evaluation of their strengths and limitations.
    \item Developed the SkeySpot Interactive Interface Toolkit, a real-time web-based tool for small and medium-sized enterprises in the construction industry, automating service key spotting, detection summaries, and analysis to enhance decision-making and operational efficiency with an end-to-end scalable solution.
\end{enumerate}

\section{Materials and methods}


This section outlines the methodology used to create the dataset, design the model architecture, and train the deep learning model for detecting service keys in digitised electrical layout plans. An overview of the development and deployment framework for SkeySpot is illustrated in Figure \ref{fig:Pipeline Diagram}.

\subsection{Dataset Preparation}
\begin{figure}[!htbp]
  \centering
  \includegraphics[width=\columnwidth]{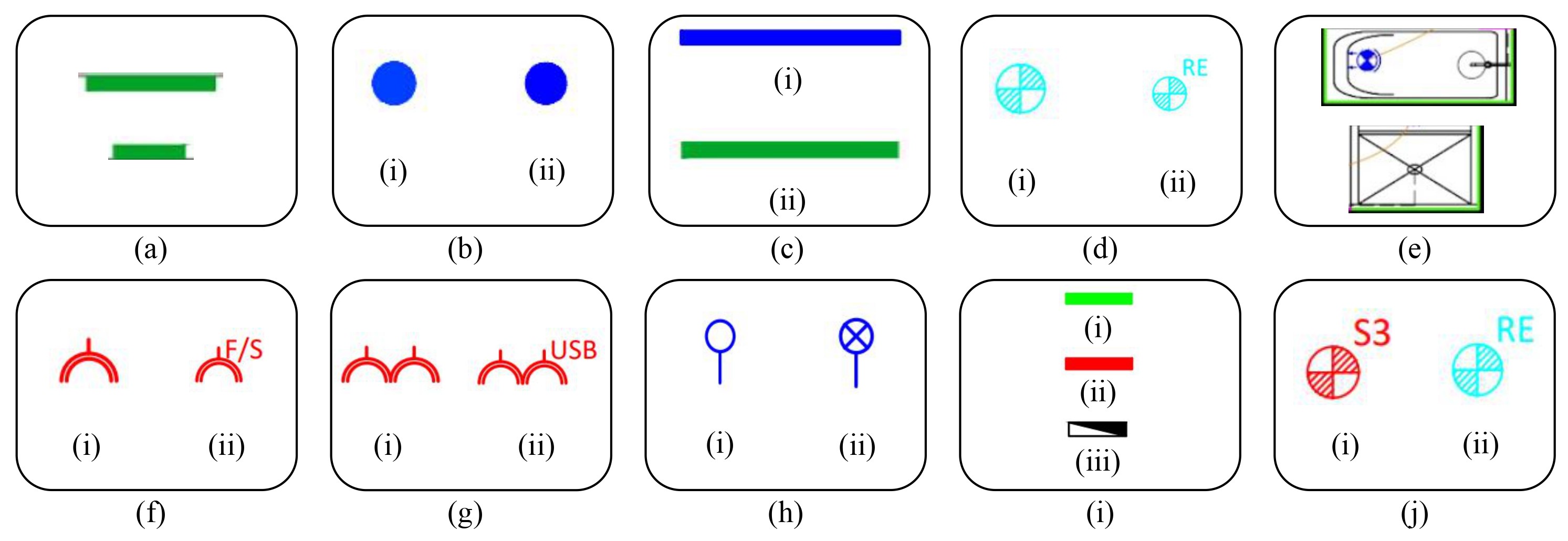}
  \caption{Cases from DELP dataset with intra-class dissimilarities or inter-class similarities that make this dataset relevant for building robust models for the application a) Radiator (b)(i) Water Entry Position (ii) Low Energy Pendant Light (c)(i) Twin LED Strip Light (ii) Radiator (d)(i) Ceiling Fan With Light Switch (ii) Recirculating Extractor Fan (e) Full Height Tiling (f)(i) Single Socket (ii) Fused Spur (g)(i) Double Socket (ii) USB Double Socket (h)(i) Internal Wall Light (ii) External Wall Light (i)(i) Gas Meter Box (ii) Electric Meter Box (iii) Consumer Unit (j)(i) Ceiling Fan With Boost Switch (ii) Recirculating Extractor Fan.}
\label{fig:variations}
\end{figure}

The Digital Electrical Layout Plans (DELP) dataset comprises 45 curated floor plans annotated with 34 distinct classes of service keys relevant to electrical systems. This dataset presents numerous challenges for automated symbol detection due to factors such as occlusion, clutter, color variations, and textual elements. The floor plans were obtained from UK-based SMEs engaged in housing estate development. Representative examples of symbol variability are shown in \autoref{fig:variations}, which includes radiators with differing shapes and varying water entry positions, as well as low-energy pendant lights that, while similar in shape, differ in color. These intra-class variations and inter-class similarities add to the complexity of the classification task, particularly in instances where symbols overlap spatially.


Each image in the dataset includes labeled annotations indicating the location and class of the service keys. A complete list of service key classes is provided in \autoref{tab:service_key}. During dataset preparation, four classes (Splash-back Tiling, Half Height Tiling, Underfloor Heating, and Cornice) with only a single instance were excluded to avoid extreme class imbalance. Annotations were created using the LabelImg tool \cite{tool}. Class-wise instance frequencies across the annotated dataset are summarised in \autoref{tab:classwise-res}.

\begin{table}[htbp!]
\centering
\caption{Service key classes with graphical symbols.}
\label{tab:service_key}
\begin{tabular}{|>{\centering\arraybackslash}m{2.9cm}|m{0.625cm}|>{\centering\arraybackslash}m{2.9cm}|m{0.625cm}|}
\hline
\textbf{Service Key} & \textbf{Icon} & \textbf{Service Key} & \textbf{Icon} \\
\hline
BT Entry Point & \parbox[c][0.625cm][c]{0.625cm}{\centering \includegraphics[width=0.625cm,height=0.625cm,keepaspectratio]{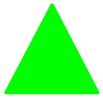}} & 
Cat 6 Data Socket & \parbox[c][0.625cm][c]{0.625cm}{\centering \includegraphics[width=0.625cm,height=0.625cm,keepaspectratio]{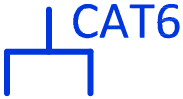}} \\
\hline
Ceiling Fan With Light Switch & \parbox[c][0.625cm][c]{0.625cm}{\centering \includegraphics[width=0.625cm,height=0.625cm,keepaspectratio]{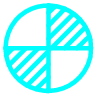}} & 
Ceiling Fan With Boost Switch & \parbox[c][0.625cm][c]{0.625cm}{\centering \includegraphics[width=0.625cm,height=0.625cm,keepaspectratio]{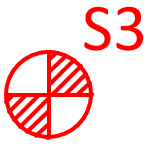}} \\
\hline
Co-Ax TV Socket & \parbox[c][0.625cm][c]{0.625cm}{\centering \includegraphics[width=0.625cm,height=0.625cm,keepaspectratio]{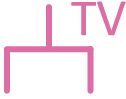}} & 
Consumer Unit & \parbox[c][0.625cm][c]{0.625cm}{\centering \includegraphics[width=0.625cm,height=0.625cm,keepaspectratio]{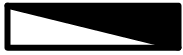}} \\
\hline
Double Socket & \parbox[c][0.625cm][c]{0.625cm}{\centering \includegraphics[width=0.625cm,height=0.625cm,keepaspectratio]{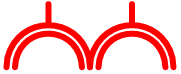}} & 
Electric Meter Box & \parbox[c][0.625cm][c]{0.625cm}{\centering \includegraphics[width=0.625cm,height=0.625cm,keepaspectratio]{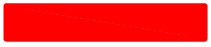}} \\
\hline
External Wall Light & \parbox[c][0.625cm][c]{0.625cm}{\centering \includegraphics[width=0.625cm,height=0.625cm,keepaspectratio]{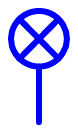}} & 
Fused Spur & \parbox[c][0.625cm][c]{0.625cm}{\centering \includegraphics[width=0.625cm,height=0.625cm,keepaspectratio]{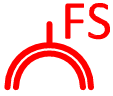}} \\
\hline
Full Height Tiling & \parbox[c][0.625cm][c]{0.625cm}{\centering \includegraphics[width=0.625cm,height=0.625cm,keepaspectratio]{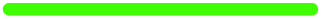}} & 
Gas Meter Box & \parbox[c][0.625cm][c]{0.625cm}{\centering \includegraphics[width=0.625cm,height=0.625cm,keepaspectratio]{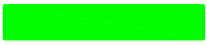}} \\
\hline
Grid Switch & \parbox[c][0.625cm][c]{0.625cm}{\centering \includegraphics[width=0.625cm,height=0.625cm,keepaspectratio]{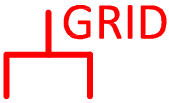}} & 
Hob Switch & \parbox[c][0.625cm][c]{0.625cm}{\centering \includegraphics[width=0.625cm,height=0.625cm,keepaspectratio]{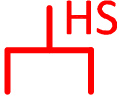}} \\
\hline
Internal Wall Light & \parbox[c][0.625cm][c]{0.625cm}{\centering \includegraphics[width=0.625cm,height=0.625cm,keepaspectratio]{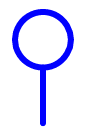}} & 
Light Switch & \parbox[c][0.625cm][c]{0.625cm}{\centering \includegraphics[width=0.625cm,height=0.625cm,keepaspectratio]{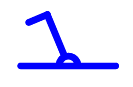}} \\
\hline
Low Energy Downlighter & \parbox[c][0.625cm][c]{0.625cm}{\centering \includegraphics[width=0.625cm,height=0.625cm,keepaspectratio]{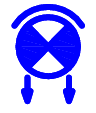}} & 
Low Energy Pendant Light & \parbox[c][0.625cm][c]{0.625cm}{\centering \includegraphics[width=0.625cm,height=0.625cm,keepaspectratio]{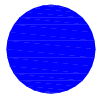}} \\
\hline
Mains Wired Smoke Detector & \parbox[c][0.625cm][c]{0.625cm}{\centering \includegraphics[width=0.625cm,height=0.625cm,keepaspectratio]{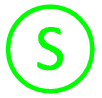}} & 
Outside Socket & \parbox[c][0.625cm][c]{0.625cm}{\centering \includegraphics[width=0.625cm,height=0.625cm,keepaspectratio]{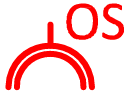}} \\
\hline
Outside Tap & \parbox[c][0.625cm][c]{0.625cm}{\centering \includegraphics[width=0.625cm,height=0.625cm,keepaspectratio]{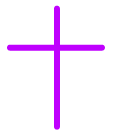}} & 
Oven Switch & \parbox[c][0.625cm][c]{0.625cm}{\centering \includegraphics[width=0.625cm,height=0.625cm,keepaspectratio]{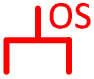}} \\
\hline
Programmable Room Thermostat & \parbox[c][0.625cm][c]{0.625cm}{\centering \includegraphics[width=0.625cm,height=0.625cm,keepaspectratio]{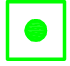}} & 
Radiator & \parbox[c][0.625cm][c]{0.625cm}{\centering \includegraphics[width=0.625cm,height=0.625cm,keepaspectratio]{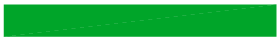}} \\
\hline
Recirculating Extractor Fan & \parbox[c][0.625cm][c]{0.625cm}{\centering \includegraphics[width=0.625cm,height=0.625cm,keepaspectratio]{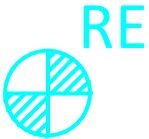}} & 
Shaver Socket & \parbox[c][0.625cm][c]{0.625cm}{\centering \includegraphics[width=0.625cm,height=0.625cm,keepaspectratio]{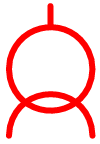}} \\
\hline
Single Socket & \parbox[c][0.625cm][c]{0.625cm}{\centering \includegraphics[width=0.625cm,height=0.625cm,keepaspectratio]{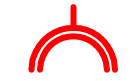}} & 
TV - Satellite Multisocket & \parbox[c][0.625cm][c]{0.625cm}{\centering \includegraphics[width=0.625cm,height=0.625cm,keepaspectratio]{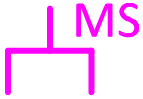}} \\
\hline
Telephone Socket & \parbox[c][0.625cm][c]{0.625cm}{\centering \includegraphics[width=0.625cm,height=0.625cm,keepaspectratio]{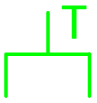}} & 
Track Light & \parbox[c][0.625cm][c]{0.625cm}{\centering \includegraphics[width=0.625cm,height=0.625cm,keepaspectratio]{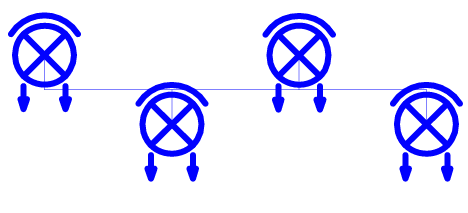}} \\
\hline
Twin LED Strip Light & \parbox[c][0.625cm][c]{0.625cm}{\centering \includegraphics[width=0.625cm,height=0.625cm,keepaspectratio]{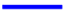}} & 
USB Double Socket & \parbox[c][0.625cm][c]{0.625cm}{\centering \includegraphics[width=0.625cm,height=0.625cm,keepaspectratio]{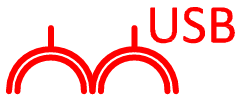}} \\
\hline
Underfloor Heating Manifold & \parbox[c][0.625cm][c]{0.625cm}{\centering \includegraphics[width=0.625cm,height=0.625cm,keepaspectratio]{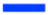}} & 
Water Entry Position & \parbox[c][0.625cm][c]{0.625cm}{\centering \includegraphics[width=0.625cm,height=0.625cm,keepaspectratio]{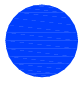}} \\
\hline
Cornice & \parbox[c][0.625cm][c]{0.625cm}{\centering \includegraphics[width=0.625cm,height=0.625cm,keepaspectratio]{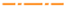}} & 
Half Height Tiling & \parbox[c][0.625cm][c]{0.625cm}{\centering \includegraphics[width=0.625cm,height=0.625cm,keepaspectratio]{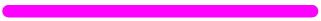}} \\
\hline
Splashback Tiling & \parbox[c][0.625cm][c]{0.625cm}{\centering \includegraphics[width=0.625cm,height=0.625cm,keepaspectratio]{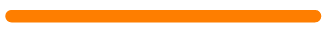}} & 
Underfloor Heating & \parbox[c][0.625cm][c]{0.625cm}{\centering \includegraphics[width=0.625cm,height=0.625cm,keepaspectratio]{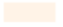}} \\
\hline
\end{tabular}
\end{table}

\subsection{Data Augmentation and Preprocessing}
\label{augmen}

To enhance the DELP dataset and ensure robust training of the pretrained models, we utilised image augmentation via the albumentations library \cite{2018arXiv180906839B}. Four augmentations were applied to the original images: 1) Vertical and horizontal flip, 2) Rotation, 3) Random crop, and 4) Brightness adjustment. These techniques enhance dataset diversity, improving model performance by enabling better generalisation across varying scenarios, which boosts accuracy and robustness.

\subsection{Model Architecture and Training}


After dataset preparation and augmentation, the pretrained models were finetuned on the DELP dataset to detect service keys. The trained model when deployed accepts an input image of the floor plan and generates bounding box coordinates detecting the service key symbols and identifies them. This information can be further processed or analysed in various applications such as automated floor analysis, building information modeling (BIM), renovation of the houses or property management.

This work leveraged two pretrained models: YOLO (You Look Only Once) \cite{yolov8} and Faster-RCNN \cite{ren2015faster}, both of which were trained on the COCO dataset \cite{lin2014microsoft}. YOLOv8’s architecture comprises three main components: Backbone, Neck, and Head. The Backbone employs a custom CSPDarknet53 with CSP connections, depth-wise separable convolutions, and residual links to efficiently extract multi-scale features. The Neck integrates an enhanced PANet and the novel C2f module, replacing YOLOv5’s C3 module, to improve feature fusion and small object detection. The anchor-free Head directly predicts object centers, streamlining non-maximum suppression (NMS) and eliminating anchor tuning, enabling real-time performance without compromising accuracy.

Faster R-CNN uses a two-stage design combining a region proposal network (RPN) with a deep CNN backbone like ResNet or VGG for feature extraction. The RPN predicts bounding boxes and objectness scores, refining proposals through RoI pooling into fixed-size feature maps for classification and regression. This anchor-based, region-specific approach enhances accuracy but increases computational complexity, contrasting with YOLOv8’s single-shot, real-time design.
YOLOv8 was fine-tuned over 200 epochs, while Faster R-CNN was trained for 4,000 iterations, with both models undergoing 5-fold cross-validation (80-20 split for training and validation) for hyperparameter tuning. Learning rates and batch sizes were optimised, and model weights were saved for each fold to support further evaluation.

\section{Results}


Following the evaluation protocol adopted by \cite{9151066}, the model performance is reported using standard object detection metrics: mAP@50, mAP@75, and the averaged mAP@[50:95], which collectively assess both localization and classification precision across varying IoU thresholds. Here mean Average Precision (mAP) measures detection accuracy across classes, while Intersection over Union (IoU) quantifies the overlap between predicted and ground truth bounding boxes. These metrics were computed on an unseen test set of 13 floor plans, with average performance across all folds was used to provide a reliable evaluation of the models' detection capabilities.

\begin{table}[!htbp]
\caption{Comparison of Faster R-CNN and YOLOv8 on 13 images with 598 instances of test set. The metrics are the mean of the mAP scores of the models trained with 5-fold cross validation.}
\label{tab:results}
\centering
\begin{tabular}{@{}lccc@{}}
\toprule
\textbf{Framework} & \textbf{mAP50 (\%)} & \textbf{mAP75 (\%)} & \textbf{mAP50-95 (\%)} \\ \midrule
Faster R-CNN       & 48.62               & 25.54               & 26.13                  \\
YOLOv8             & 81.12               & 46.86               & 48.24                  \\ \bottomrule
\end{tabular}
\end{table}

\begin{table*}[!htbp]
\centering
\caption{Performance metrics for different service keys on test set for the best YOLOv8 model from 5-fold cross validation}
\label{tab:classwise-res}
\begin{tabular}{@{}cp{4cm}ccccc@{}}
\toprule
\textbf{Service ID} & \textbf{Service Keys}                                                            & \textbf{Frequency in Train Set} & \textbf{Frequency in Test Set} & \textbf{mAP50 (\%)} & \textbf{mAP75 (\%)} & \textbf{mAP50-95 (\%)} \\  \midrule

0                   & BT Entry Point                                                                   & 14                 & 4                 & 65.34               & 0                   & 25.42                  \\
1                   & Cat 6 Data Socket                                                                & 73                 & 18                & 93.98               & 69.57               & 59.44                  \\
2                   & Ceiling Fan With Light Switch & 48                 & 24                & 93.66               & 30.12               & 44.66                  \\
3                   & Ceiling Fan With Boost Switch                   & 23                 & 0                 & -                   & -                   & -                      \\
4                   & Co-Ax TV Socket                                                                  & 42                 & 10                & 83.40                & 47.31               & 54.90                   \\
5                   & Consumer Unit                                                                    & 13                 & 4                 & 75.00                  & 43.50                & 43.90                   \\
6                   & Double Socket                                                                    & 348                & 121               & 97.00                  & 60.68               & 60.44                  \\
7                   & Electric Meter Box                                                               & 16                 & 5                 & 89.88               & 78.49               & 58.74                  \\
8                   & External Wall Light                                                              & 32                 & 10                & 99.50                & 73.38               & 65.38                  \\
9                   & Fused Spur                                                                       & 5                  & 0                 & -                   & -                   & -                      \\
10                  & Full Height Tiling                                                               & 31                 & 14                & 95.32               & 93.98               & 84.36                  \\
11                  & Gas Meter Box                                                                    & 15                 & 4                 & 82.08               & 52.64               & 46.54                  \\
12                  & Grid Switch                                                                      & 9                  & 5                 & 56.44               & 0                   & 16.34                  \\
13                  & Hob Switch                                                                       & 17                 & 4                 & 68.62               & 36.77               & 33.04                  \\
14                  & Internal Wall Light                                                              & 6                  & 1                 & 99.50                & 0                   & 47.72                  \\
15                  & Light Switch                                                                     & 218                & 78                & 91.96               & 36.61               & 46.38                  \\
16                  & Low Energy Downlighter                                                           & 282                & 87                & 98.66               & 94.86               & 75.64                  \\
17                  & Low Energy Pendant Light                                                         & 151                & 46                & 99.50                & 99.50                & 83.82                  \\
18                  & Mains Wired Smoke Detector                                                       & 42                 & 17                & 99.50                & 81.83               & 70.12                  \\
19                  & Outside Socket                                                                   & 13                 & 6                 & 83.00                  & 63.60                & 47.68                  \\
20                  & Outside Tap                                                                      & 14                 & 5                 & 93.70                & 90.57               & 67.96                  \\
21                  & Oven Switch                                                                      & 19                 & 4                 & 65.80                & 7.85                & 24.12                  \\
22                  & Programmable Room Thermostat                                                     & 34                 & 10                & 91.94               & 25.66               & 35.78                  \\
23                  & Radiator                                                                         & 162                & 67                & 94.30                & 62.50                & 59.16                  \\
24                  & Recirculating Extractor Fan                                                      & 14                 & 0                 & -                   & -                   & -                      \\
25                  & Shaver Socket                                                                    & 33                 & 9                 & 99.50                & 61.07               & 62.78                  \\
26                  & Single Socket                                                                    & 33                 & 10                & 92.72               & 28.30                & 48.36                  \\
27                  & TV - Satellite Multisocket                                                       & 2                  & 6                 & 24.22               & 21.60                & 19.5                   \\
28                  & Telephone Socket                                                                 & 34                 & 13                & 71.80                & 25.31               & 34.56                  \\
29                  & Track Light                                                                      & 3                  & 2                 & 21.00                  & 0                   & 8.40                    \\
30                  & Twin LED Strip Light                                                             & 9                  & 0                 & -                   & -                   & -                      \\
31                  & USB Double Socket                                                                & 78                 & 5                 & 43.02               & 20.87               & 26.26                  \\
32                  & Underfloor Heating Manifold                                                      & 11                 & 0                 & -                   & -                   & -                      \\
33                  & Water Entry Position                                                             & 12                 & 5                 & 80.00                  & 52.26               & 47.26                  \\ \bottomrule
\end{tabular}
\end{table*}


The performance and accuracy of the finetuned faster R-CNN and YOLOv8 models for detecting service keys in DELP are analysed and compared upon multiple metrics. The evaluation was conducted using the the 13 hold-out test floor plan images, and the results of each model's performance are summarised in Table \ref{tab:results}. The class-wise performance metrics for the service keys are presented in the \ref{tab:classwise-res}. This table also highlights the differences in the number of instances between the training and test datasets for each service key. Notably, some service keys are marked with a hyphen, indicating the absence of corresponding instances in the test dataset.

\section{SkeySpot Toolkit}
\subsection{Features}
The SkeySpot toolkit offers an intuitive interface for digital annotation and analysis, allowing users to generate bounding boxes around detected labels, using color-coded service IDs for easy differentiation. Skeyspot generates annotated layouts, providing detailed detection insights for other downstream decision making tasks. The tool supports visualization for each service key, quantifies them by class and count. For large scale analysis involving multiple floor plans, users can employ the batch processing mode. The tool enables users to fine-tune confidence scores, download annotated images, and generate layout-wise detection summaries, and could offer SMEs a practical alternative to traditional workflows.

\subsection{Deployment}
SkeySpot toolkit was deployed using Streamlit Cloud, an open-source Python framework that enables easy deployment and sharing of web-based applications. Streamlit allows for automatic scaling and monitoring of the deployed application, ensuring that it is always available, responsive and can be used by anyone with an internet connection.
While the initial training leveraged both Faster R-CNN and YOLOv8 models, the final deployment on the Streamlit platform exclusively uses the best model: YOLOv8. The Streamlit web app uses maximum of 2 CPU cores, 2.7 GB of memory, and supports upto 50 GB of storage, which are adequate resources for running SkeySpot toolkit effectively. The SkeySpot toolkit is publicly accessible at \url{https://skeyspot.streamlit.app}. In addition, all source code and dataset details are available at \url{https://github.com/HAIx-Lab/Skeyspot} to ensure reproducibility. 

\begin{figure*}[htbp]
  \centering
  \includegraphics[width=.85\textwidth]{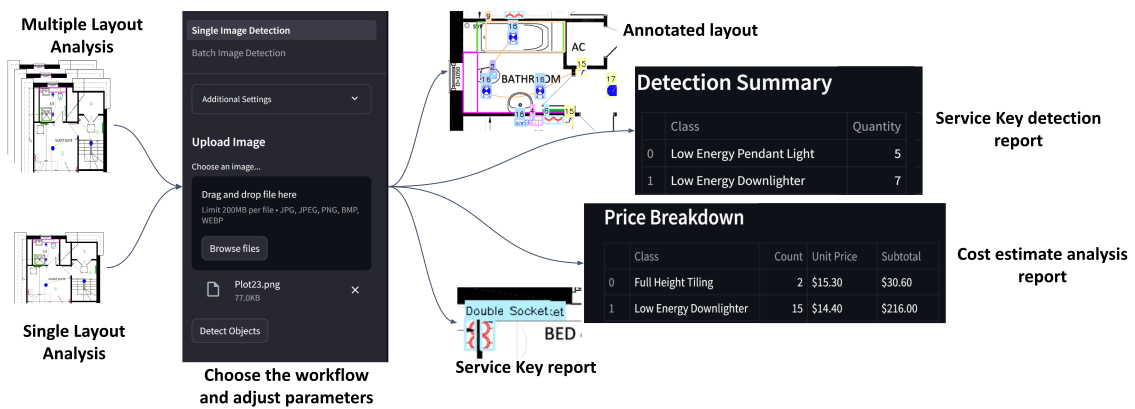}
  \caption{Workflow and features of the SkeySpot: User lands on the SkeySpot app page, uploads single or multiple floor plans. The reports include annotated images with service IDs, total count and customisable cost estimate reports of service keys in the image(s). Users can visualise individual service keys. The reports are packaged into a zip file, available for download.}
  \label{fig:user_workflow}
\end{figure*}

\begin{figure*}[htbp]
  \centering
  \includegraphics[width=.80\textwidth]{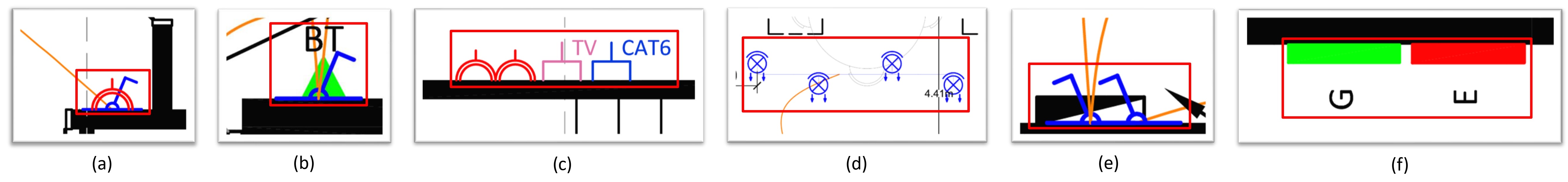}
  \caption{Detection results for service keys in challenging scenarios. (a) Light switch and single socket both are detected even though they overlap. (b) BT entry point detected but light switch is missed in the occlusion. (c) All three service keys identified despite close placements. (d) Track light misclassified as low-energy downlighter. (e) Model confused by overlaps; consumer unit undetected, light switch identified. (f) Electric meter box detected; gas meter box misclassified as radiator.}
  \label{fig:detection_results}
\end{figure*}

\subsection{Interactive Interface}
User workflow on SkeySpot toolkit is represented in the \autoref{fig:user_workflow}. The SkeySpot toolkit is designed with two interactive pages to enhance user experience. The first page allows users to upload a single floor plan and assess the results with a focus on visualizing specific service keys in isolation, quantifying them by class and total count. This feature helps users fine-tune parameters for their data to align with their expected outcomes. The second page supports batch processing of multiple floor plans, incorporating cost analysis features. Users often require details about specific service keys across housing estates for tasks like cost estimation, maintenance, and retrofitting. To facilitate this, users can input per-piece rates for selected service keys, leading to a thorough cost breakdown for all uploaded floor plans. Additionally, both pages offer the option to download a zip file that contains annotated images and a summary of image-wise detections, ensuring easy access to important information.



\section{Discussion}

The YOLOv8 model demonstrated superior performance with an mAP50 score of 81.1\%, significantly outperforming Faster R-CNN, which achieved a score of 48.6\%. This performance disparity was primarily attributed to Faster R-CNN's challenges in handling densely clustered or overlapping service keys and its frequent generation of redundant bounding boxes for the same instance.

YOLOv8 exhibited notably better detection accuracy in scenarios involving crowded or overlapping symbols, as illustrated in Figure~\ref{fig:detection_results}. The model achieved robust results for service keys with higher representation in the training data, such as double sockets, light switches, and low-energy pendant lights. Performance declined for under-represented classes, such as track lights, highlighting the importance of instance frequency. Interestingly, service keys like internal and external wall lights demonstrated high mAP50 scores despite their limited training data, suggesting that distinct shape features can compensate for sample scarcity. In contrast, service keys with minimal visual differentiation—e.g., USB double sockets, grid switches, and hob switches—tended to underperform due to high intra-class similarity and reliance on textual distinctions. Conversely, keys with visually salient or unique structural patterns were more reliably detected.

From a broader perspective, these results indicate the potential of deep learning-based symbol spotting systems to support scalable digitisation and interpretation of architectural floor plans. While this work focuses on electrical layouts for its mere variety in symbols, such tools lay the foundation for automating documentation tasks traditionally requiring significant manual labour \cite{pizarro2022automatic}. Structured symbol detection may serve as a precursor for downstream tasks such as lighting zoning, load distribution estimation, and HVAC planning—critical inputs for energy simulation engines \cite{bonomolo2021building}. Although SkeySpot does not directly perform these functions, the standardised and machine-readable outputs produced through detection offer interoperability with tools in BIM, thereby enabling future integration.

Moreover, by lowering the technical entry barrier for non-CAD users through a lightweight tool like SkeySpot, this work demonstrates how AI can democratize access to essential information embedded in legacy building documents. This is particularly valuable for small and medium enterprises (SMEs) that lack access to expensive CAD software or skilled personnel. Consequently, this facilitates inclusive participation in sustainable building initiatives and compliance with energy codes.

Automated service key detection faces notable limitations, including limited annotated data, significant variation in electrical layout symbols across companies hindering standardisation, and a high initial setup effort for dataset curation and annotation alignment. Therefore, improvements could involve incorporating synthetic floor plan generation or advanced data augmentation techniques (e.g., mixup, cutout, grid masking), with appropriate safeguards to preserve label integrity \cite{hong2021synthetic}. Data augmentation could also enable the use of transformers or attention-based mechanisms for scene segmentation tasks. These refinements could extend the applicability of such systems to broader domains, including policy compliance automation, architectural innovation, and sustainability-driven design workflows, echoing emerging directions in AI-assisted urban and infrastructure planning. 

\section{Conclusion}
This work introduces the first curated dataset-DELP for annotated service key detection in digitised electrical floor plans. Several object detection models are benchmarked, with YOLOv8 identified as the top-performing framework, achieving an mAP@50 of 82.5\% and an average mAP of 81.1\%. SkeySpot, the first real-time web-based platform for automatic service key detection in electrical layouts, is developed. Together, DELP dataset and SkeySpot offer a novel and scalable automatic electrical symbol detection workflow that streamlines analysis and supports downstream decision-making for property developers, architects, and maintenance agencies.

\noindent{Acknowledgment:}
We thank Christopher Theobald, Brad Davies, and Mersea Homes Ltd for generously providing access to the dataset used in this study and thank to the IIT Gandhinagar for IRG grant - IP/IITGN/CSE/YM/2324/05.





\bibliography{references}

\end{document}